\documentclass[preprint,12pt]{elsarticle}

\usepackage{graphicx}
\usepackage{amssymb}
\usepackage{multirow}
\usepackage{booktabs}
\usepackage{amssymb}
\usepackage{mathrsfs}
\usepackage{longtable}
\usepackage{lscape}
\usepackage{tabularx}
\usepackage{epsfig}
\usepackage{colortbl}
\usepackage{amsthm}
\usepackage{amsmath}
\usepackage{mathtools}
\newdefinition{definition}{Definition}
\newdefinition{example}{Example}
\newdefinition{remark}{Remark}
\newtheorem{theorem}{Theorem}[section]

\biboptions{square,sort&compress}

\journal{XXXX}

\begin{document}

\begin{frontmatter}



\title{Belief and plausibility measures for D numbers}

\author[NWPU]{Xinyang Deng\corref{COR}}
\ead{xinyang.deng@nwpu.edu.cn}

\cortext[COR]{Corresponding author.}

\address[NWPU]{School of Electronics and Information, Northwestern Polytechnical University, Xi'an 710072, China}

\begin{abstract}
As a generalization of Dempster-Shafer theory, D number theory provides a framework to deal with uncertain information with non-exclusiveness and incompleteness. However, some basic concepts in D number theory are not well defined. In this note, the belief and plausibility measures for D numbers have been proposed, and basic properties of these measures have been revealed as well.
\end{abstract}

\end{frontmatter}

\section{Introduction}
Dempster-Shafer evidence theory (DST) \cite{Dempster1967,shafer1976mathematical} is one of the most popular theories for dealing with uncertain information, and has been widely used in various fields \cite{liu2017combination,deng2019zero,yager2017maxitive254}. But it is limited by some hypotheses and constraints that are hardly satisfied in some situation \cite{deng2012d,deng2014supplier411,xydeng2017DNCR,deng2019evaluating}. There are two main aspects.  First, in DST a frame of discernment (FOD) must be composed of mutually exclusive elements, which is called the FOD's exclusiveness hypothesis. Second, in DST the sum of basic probabilities or mass $m(.)$ in a basic probability assignment (BPA) must be 1, which is called the BPA's completeness constraint.

To overcome the above-mentioned limitations in DST, a new framework called D number theory (DNT) has been proposed \cite{deng2012d,xydeng2017DNCR,deng2019d} for the fusion of uncertain information with non-exclusiveness and incompleteness. This is a developing theory. In previous studies, the definition of D numbers, combination rule and uncertainty measure for D numbers have been given one after another \cite{deng2012d,xydeng2017DNCR,deng2019d,IJISTUDNumbers}. At the same time, a belief measure and a plausibility measure for D numbers were developed \cite{deng2019d,IJISTUDNumbers}. However, after deep and further research, it was found that the previous developed belief and plausibility measures of D numbers are not satisfactory. They do not satisfy the property of $Bel(A) + Pl(\bar A) = 1$ for any subset $A$ belonging to the FOD. Facing that, this note gives a new pair of belief and the plausibility measures, and the properties of the new belief and plausibility measures are presented.

The rest of this note is organized as follows. Section \ref{SectBasicsDST} gives a brief introduction about DST and DNT. In Section \ref{SectionProposed}, new belief measure and plausibility measure for D numbers are proposed and basic properties of these measures are presented. Finally, Section \ref{SectConclusion} concludes the note.

\section{Basics of Dempster-Shafer theory and D number theory}\label{SectBasicsDST}
In this section, some basic definitions and concepts about DST and DNT are given as below.

Let $\Omega$ be a set of $N$ mutually exclusive and collectively exhaustive events, indicated by
\begin{equation}
\Omega  = \{ q_1 ,q_2 , \cdots ,q_i , \cdots ,q_N \}
\end{equation}
where set $\Omega$ is called a frame of discernment (FOD). The power set of $\Omega$ is indicated by $2^\Omega$, namely
\begin{equation}
2^\Omega   = \{ \emptyset ,\{ q_1 \} , \cdots ,\{ q_N \} ,\{ q_1
,q_2 \} , \cdots ,\{ q_1 ,q_2 , \cdots ,q_i \} , \cdots ,\Omega \}.
\end{equation}
The elements of $2^\Omega$ or subsets of $\Omega$ are called propositions.

\begin{definition}
Let a FOD be $\Omega = \{ q_1 ,q_2 , \cdots, q_N \}$, a mass function defined on $\Omega$ is a mapping $m$ from  $2^\Omega$ to $[0,1]$, formally defined by:
\begin{equation}
m: \quad 2^\Omega \to [0,1]
\end{equation}
which satisfies the following condition:
\begin{eqnarray}
m(\emptyset ) = 0 \quad {\rm{and}} \quad \sum\limits_{A \subseteq \Omega }{m(A) = 1}.
\end{eqnarray}
\end{definition}

Given a BPA, its associated belief measure $Bel_{m}$ and plausibility measure $Pl_{m}$ express the lower bound and upper bound of the support degree to each proposition in that BPA, respectively. They are defined as
\begin{equation}
Bel_{m} (A) = \sum\limits_{B \subseteq A} {m(B)},
\end{equation}
\begin{equation}
Pl_{m}(A) = \sum\limits_{B \cap A \ne \emptyset }{m(B)},
\end{equation}
Obviously, $Pl_{m}(A) \ge Bel_{m}(A)$ for each $A \subseteq \Omega$, and $[Bel_{m}(A), Pl_{m}(A)]$ is called the belief interval of $A$ in $m$.

D number theory (DNT) is a new theoretical framework for uncertainty reasoning that has generalized DST to the situation of non-exclusive and incomplete information.

\begin{definition}\label{DefDNumbers}
Let $\Theta$ be a nonempty finite set $\Theta  = \{ \theta_1 ,\theta_2 , \cdots ,\theta_N \}$, a D number is a mapping formulated by
\begin{equation}
D: 2^{\Theta} \to [0,1]
\end{equation}
with
\begin{eqnarray}
\sum\limits_{B \subseteq \Theta } {D(B)} \le 1  \quad {\rm {and}} \quad
D(\emptyset ) = 0
\end{eqnarray}
where $\emptyset$ is the empty set and $B$ is a subset of $\Theta$.
\end{definition}

In DNT, the elements in FOD $\Theta$ are not required to be mutually exclusive. Regarding the non-exclusiveness in DNT, a membership function is developed to measure the non-exclusive degrees between elements in $\Theta$.
\begin{definition}\label{DefNonExclusiveness}
Given $B_i ,B_j  \in 2^{\Theta}$, the non-exclusive degree between $B_i$ and $B_j$ is characterized by a mapping $u$:
\begin{equation}
u :2^{\Theta}   \times 2^{\Theta}   \to [0,1]
\end{equation}
with
\begin{equation}
u (B_i ,B_j ) = \left\{ \begin{array}{l}
 1,\quad B_i  \cap B_j  \ne \emptyset   \\
 p,\quad B_i  \cap B_j  = \emptyset   \\
 \end{array} \right.
\end{equation}
and
\begin{equation}
u (B_i ,B_j ) = u (B_j ,B_i )
\end{equation}
where $0 \le p \le 1$. All non-exclusive degrees $u (B_i ,B_j )$, $B_i ,B_j  \in 2^{\Theta}$, make up a matrix ${\bf{U}}$.
\end{definition}

If $\sum\limits_{B \subseteq \Theta } {D(B) = 1}$, the corresponding D number is information-complete. By contrast, if $\sum\limits_{B \subseteq \Theta } {D(B) < 1}$ the D number is information-incomplete. In this note, we focus on the case of complete information, but the proposed measures can be naturally generalized to the case of information-incomplete D numbers.

\section{Proposed belief and plausibility measures for D numbers}\label{SectionProposed}
In this section, a new pair of belief measure and plausibility measure for D numbers is proposed as follows.

\begin{definition}
Let $D$ represent a D number defined on $\Theta$, the belief measure for $D$ is mapping
\[
Bel: 2^{\Theta}   \to [0,1]
\]
satisfying
\begin{equation}\label{EqBelMeasureDNT}
Bel(A) = \sum\limits_{B \subseteq A} {D(B)\left[ {1 - u(B,\bar A)} \right]}
\end{equation}
for any $A \subseteq \Theta$. Since ${u(B,\bar A)} = 1$ for any $B \not\subset A$ in terms of the definition of non-exclusive degree $u$, Eq. (\ref{EqBelMeasureDNT}) can be written as
\begin{equation}\label{EqBelMeasure}
Bel(A) = \sum\limits_{B \subseteq \Theta} {D(B)\left[ {1 - u(B,\bar A)} \right]}
\end{equation}
\end{definition}

\begin{definition}
Let $D$ represent a D number defined on $\Theta$, the plausibility measure for $D$ is mapping
\[
Pl: 2^{\Theta}  \to [0,1]
\]
satisfying
\begin{equation}
Pl(A) = \sum\limits_{B \cap A \ne \emptyset } {D(B)}  + \sum\limits_{B \cap A = \emptyset } {u(B,A)D(B)}
\end{equation}
where $A, B \subseteq \Theta$. Because $u(B ,A) = 1$ for $B  \cap A  \ne \emptyset$ in terms of the definition of non-exclusive degree $u$, the plausibility measure $Pl$ can also be written as
\begin{equation}\label{EqPlMeasureDNT}
Pl(A) = \sum\limits_{B \subseteq \Theta} {u(B,A) D(B)}
\end{equation}
\end{definition}

It can be proved that the proposed measures satisfies the following basic properties.

\begin{theorem}
\[Pl(A) \ge Bel(A) \]
\end{theorem}

\begin{theorem}
\[
Bel(A) + Bel(\bar A) \le 1
\]
\end{theorem}

\begin{theorem}
\[
Pl(A) + Pl(\bar A) \ge 1
\]
\end{theorem}

%\begin{theorem}
%If $A \subseteq B$, then $Bel(A) \le Pl(B)$ and $Pl(A) \le Pl(B)$
%\end{theorem}

\begin{theorem}
\[Bel(A) + Pl(\bar A) = 1\]
\end{theorem}

If letting ${\bf{Pl}}$ be the vector form of plausibility measure, and {\bf{D}} the vector form of a D number, according to Eq. (\ref{EqPlMeasureDNT}), we have the following relationship
\begin{equation}
{\bf{Pl}} = {\bf{D}} \cdot {\bf{U}}
\end{equation}

In terms of the above definitions and properties, a D number {\bf{D}}, and its belief measure ${\bf{Bel}}$ and plausibility measure ${\bf{Pl}}$ are all in one to one correspondence. As similar as DST, $[Bel(A), Pl(A)]$ forms a belief interval about the possibility of $A$ in DNT. It is easy to find that the $Bel$ and $Pl$ for D numbers will degenerate to classical belief measure and plausibility measure in DST if the associated D number is a BPA in fact.

\section{Conclusion}\label{SectConclusion}
This note has studied the issue of belief and plausibility measures in DNT. New belief and plausibility measures for D numbers have been proposed, and the properties of these measures are studied. Based on the proposed belief and plausibility measures, the framework of DNT is enriched and the application foundation of DNT is more solid.

%
%\section*{Acknowledgments}
%The work is partially supported by National Natural Science Foundation of China (Program Nos. 61703338, 61671384).



%% References


\bibliographystyle{elsarticle-num}
\bibliography{references}






% \end{thebibliography}

\end{document}